\documentclass{article}
\usepackage{xurl}
\usepackage{hyperref}
\usepackage{color,soul}
\usepackage{multirow}
\usepackage{stywhispers,amsmath,epsfig}
\usepackage{booktabs}
\usepackage{array,multirow,graphicx}
\usepackage[table]{xcolor}
\usepackage{microtype}

\newlength{\bibitemsep}
\newlength{\bibparskip}
\let\oldthebibliography\thebibliography
\renewcommand\thebibliography[1]{%
  \oldthebibliography{#1}%
  \setlength{\parskip}{\bibitemsep}%
  \setlength{\itemsep}{\bibparskip}%
}
\newcommand{\STAB}[1]{\begin{tabular}{@{}c@{}}#1\end{tabular}}
\newcommand\partialmidrule[2]{%
   \cmidrule[\heavyrulewidth](#1){#2}
   \addlinespace[-\belowrulesep]}

\title{A Vision for Cleaner Rivers: Harnessing Snapshot Hyperspectral Imaging to Detect Macro-Plastic Litter}
\name{\scalebox{0.96}{Nathaniel Hanson$^{1*\dag}$, Ahmet Demirkaya
$^{2\dag}$, Deniz Erdoğmuş$^{2}$, Aron Stubbins$^{3}$,\textit{Taşkın Padır}$^{1}$, \textit{Tales Imbiriba}$^{2}$}
\thanks{Research was sponsored by the United States Army Core of Engineers (USACE) Engineer Research and Development Center (ERDC) Geospatial Research Laboratory (GRL) and was accomplished under Cooperative Agreement Federal Award Identification Number (FAIN) W9132V-22-2-0001. The views and conclusions contained in this document are those of the authors and should not be interpreted as representing the official policies, either expressed or implied, of USACE EDRC GRL or the U.S. Government. The U.S. Government is authorized to reproduce and distribute reprints for Government purposes notwithstanding any copyright notation herein.}
\thanks{\dag Equal Contribution}
\thanks{*Corresponding author {\tt\small hanson.n@northeastern.edu}}}
\address{$^{1}$Institute for Experiential Robotics, ${^2}$ Department of Electrical and Computer Engineering,\\
${^3}$Marine Science Center, Northeastern University, Boston, MA, USA}

\begin{document}
%
\maketitle
\begin{abstract}
Plastic waste entering the riverine harms local ecosystems leading to  negative ecological and economic impacts. Large parcels of plastic waste are transported from inland to oceans leading to a global scale problem of floating debris fields. In this context, efficient and automatized monitoring of mismanaged plastic waste is paramount. To address this problem, we analyze the feasibility of macro-plastic litter detection using computational imaging approaches in river-like scenarios. We enable near-real-time tracking of partially submerged plastics by using snapshot Visible-Shortwave Infrared hyperspectral imaging.  
Our experiments indicate that imaging strategies associated with machine learning classification approaches can lead to high detection accuracy even in challenging scenarios, especially when leveraging hyperspectral data and nonlinear classifiers. All code, data, and models are available online: \url{https://github.com/RIVeR-Lab/hyperspectral_macro_plastic_detection}
\end{abstract}
\section{Introduction}
\label{sec:intro}

Mismanaged plastic waste (MPW) entering the riverine causes negative impacts on ecology endangering aquatic species. Its presence leads to substantial economic losses in the tourism industry, and added cost from increased shoreline cleaning efforts~\cite{nyberg2023leaving}. Furthermore, MPW reaching the river is associated with the spread and accumulation of plastic waste in the oceans, with conservative estimates~\cite{nyberg2023leaving} indicating that 0.8 million tonnes of MPW enter rivers annually in 2015, which affects 84 \% rivers by surface area, globally. Furthermore, different studies emphasize the knowledge gap regarding MPW that reach rivers and oceans~\cite{halle2016understanding, nyberg2023leaving}. Recent reports indicate that  most plastic waste does not reach oceans indicating that rivers can act as plastics reservoirs~\cite{van2022rivers}. However, the plastic flux and accumulation in river networks are not yet fully understood.


\begin{figure}[t]
\centering
    \includegraphics[width=1.0\linewidth]{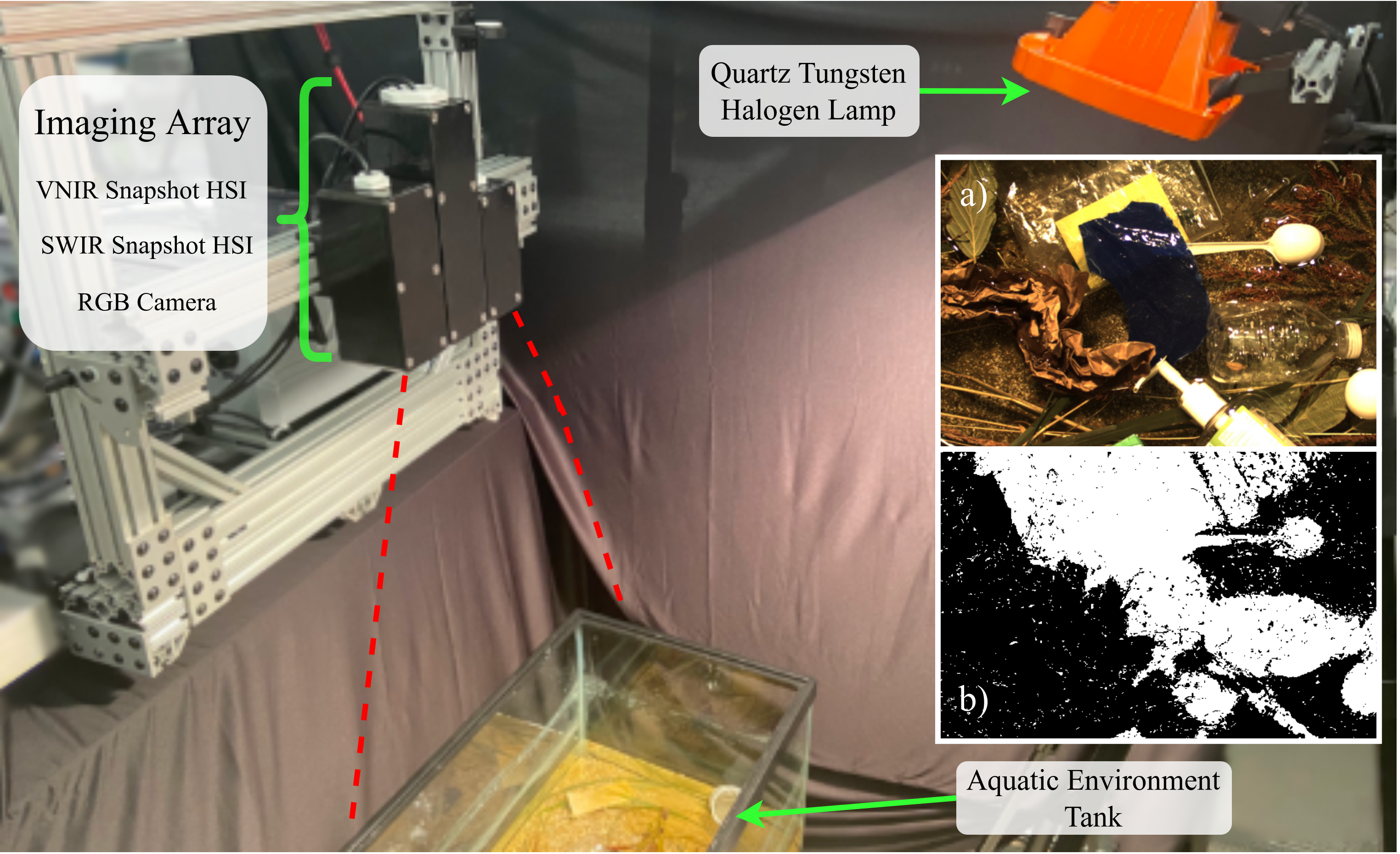}
    \caption{Experimental setup to emulate planned deployed plastic detection system above waterways. a) RGB scene registered to HSI as seen by imaging area; b) Predicted outputs on for plastic detection in cluttered aquatic environment. \textit{N.B.} White indicates positive plastic detection.}
    \label{fig:plastics_title_picture}
    \vspace{-1.5em}
\end{figure}

In this context, recent efforts within visual plastic waste monitoring have focused on developing human-assisted \cite{van2020plastic} or computer vision tools and using remote sensing strategies to detect plastic litter in water bodies~\cite{corbari2020indoor, karlsson2016hyperspectral, tasseron2022toward}.  
Most works use multispectral or pushbroom hyperspectral imaging (HSI) to better characterize plastic litter reflectances. 
Recent work analyzes the spectral reflectances of macro-plastics, both wet~\cite{moshtaghi2021spectral} and dry~\cite{corbari2020indoor} scenarios, in a controlled environment using a high-resolution point spectrometer. 
Spectral signatures of micro-plastics were also analyzed using multiple hyperspectral instruments~\cite{karlsson2016hyperspectral, garaba2018airborne}. For instance,  clustering strategies and hyperspectral imaging from multiple instruments were leveraged for micro-plastics detection in seawater filtrates~\cite{karlsson2016hyperspectral}, while~\cite{garaba2018airborne} used AVIRIS data to locate micro-plastics concentrations in the marine environment.
While micro-plastic detection and estimation are of great concern in ocean pollution, in this contribution, we focus on macro-plastic analysis and detection.

In remote-sensing and earth sciences, different spectral indexes were defined for the detection and identification of different land-cover components~\cite{calatrava2023recursive}. In~\cite{biermann2020finding} the authors propose a floating debris index (FDI) to calculate the presence of floating aggregations. However, FDI is not plastic-specific and further validation for plastic detection applications is still required. 
In~\cite{mehrubeoglu2020detection} the authors propose four indexes for plastic detection using bands in the SWIR (900-1700 nm) range.

Solutions more specific for riverine scenarios have also been explored. 
Previous works have shown promising results using Unmanned Aquatic Vehicles \cite{alboody2023new} and Unmanned Aerial Vehicles \cite{cortesi2022uav}. These systems make assumptions regarding the vehicle's positioning relative to the plastics. Such assumptions are unrealistic in flowing water, where movement of the collection target during acquisition results in severe image distortion. \cite{tasseron2021advancing}
advocated a similar transfer approach from a controlled laboratory environment to detection along a river bank but their solutions required the targets to remain in a stable position for imaging. 

In~\cite{tasseron2022toward} the authors use hyperspectral cameras to collect images with 100 equally spaced bands in the VIR-SWIR range. They use the setup to evaluate plastic classification using support-vector machines (SVMs) and Spectral Angle Mapper (SAM), the latter due to the spectral variability caused by illumination factors~\cite{borsoi2021spectral}.
This highlights another challenge in plastic detection. Many of the prior works assume excellent spectral resolution ($\leq$10 nm) uniformly
sampled across the device range. The current state-of-the-art in snapshot imaging sacrifices spectral resolution for real-time spatial acquisition.

Although many works focus on wide portions of the electromagnetic spectrum ranging from VIS to SWIR, very few works analyzing standard RGB images for MWP exist. \cite{van2020automated} uses deep learning methods and images extracted from video footage for floating plastic detection in rivers. 
Nevertheless, an understanding of the advantages of multispectral over RBG imaging for the detection of MWP is missing in the literature.

In this work, we analyze the feasibility of imaging approaches with both RGB and snapshot hyperspectral imaging paradigms for macro-plastic detection, especially in river-like scenarios.
We analyze a variety of consumer and industrial plastic types, their spectral signatures ranging from visible (VIS) to short-wave infrared (SWIR) portions of the electromagnetic spectrum. The work builds on our previous efforts to build detection pipelines for partially occluded objects using spectral signatures \cite{hanson2022occluded}. As detection strategies, we employ machine learning algorithms which are trained and tested with lab-acquired images. 
To stress the methodology to river-like scenarios we design an experiment to simulate a river with turbid and settled water and evaluate the transference of the algorithms to this different context.

\section{Methods \& Materials}
We utilize the HSI system detailed in \cite{hanson2023hyperdrive} and illustrated in Fig.~\ref{fig:plastics_title_picture}. This system includes two snapshot hyperspectral cameras: VNIR (IMEC MQ022HG-IM-SM5X5-NIR) and SWIR (IMEC SCD640-SM3X3-NIR). Together these two cameras sense the electromagnetic system from 660-1700 nm with 33 wavelength channels. The cameras are mounted on the same axis as an RGB pansharpening camera (Allied Vision Alvium). In this configuration, the cameras were pointed at $0^{\circ}$ nadir above a collection source. A full spectrum Quartz Tungsten Halogen (QTH) provided illumination emulation of the outdoor solar spectra. Integration times for the two cameras were set to 2 ms and 10 ms, respectively. It should be noted that these integration times will be shorter outdoors. Nonetheless, these acquisition times are short enough to avoid motion artifacts.

The system was reflectance calibrated using a calibrated reflectance target, 95\%  compressed PTFE. Reflectance was calculated using the following formula:

\begin{equation}
    \label{eqn:standard_reflectance}
    \text{reflectance} = \frac{\text{signal} - \text{signal}_{\text{dark}}}{\text{signal}_{\text{reference}} - \text{signal}_{\text{dark}}}
\end{equation}
In Eq.~\eqref{eqn:standard_reflectance}, the signal is the current reading from a pixel in the hyperspectral image. The dark signal is the averaged signal value when the device is covered and not collecting light. This processing step yields data properly normalized to a [0,1] range, which is preferable for classifier training.

Once the images are normalized, they are registered with a perspective transformation via a homography matrix. This step enables the overlapping field of views to be included in a common frame and upsamples the datacube to the dimension of the high-resolution RGB camera. The final spatial resolution is 1052 $\times$ 1588 pixels with 33 spectral bands.

\subsection{Experimental Scenarios}
\vspace{-0.5em}
Presently, we are interested in the application of imaging modalities (RGB and hyperspectral) and machine learning methods for MPW detection in river-like scenarios. Therefore we construct experimental scenarios to compare $(i)$ RGB and hyperspectral data suitability for the task, and $(ii)$ the capability of ML models trained in lab-controlled scenarios to generalize to river-like scenarios. 
To achieve those goals we constructed 10 different scenarios to test the increasing complexity of data. The scenario list described in Table~\ref{tab:scenario_setup} details the items present in the scene, the composition of those items, the background upon which the data were captured, and whether the set was used to train or test the various approaches. The background is important when collecting data, as it provides a non-negligible contribution to the observed spectral of transparent items. 

To ensure a diverse set of training data, we utilized a standardized set of manufacturing plastics (ICOMold), which not only have standardized shapes but also contain surface features (flashing, ridges) adding to nonuniform surface reflections. We selected the following plastics as representative of common detritus in aquatic environments: \textit{\{Polypropylene (PP), Thermoplastic Elastomers (TPE), Polymethyl Methacrylate (PMMA), Polycarbonate (PC), Polyethylene Terephthalate (PET), Acrylonitrile Styrene Acrylate (ASA), High Density Polyethylene (HDPE), Low Density Polyethylene (LDPE), Acrylonitrile Butadiene Styrene (ABS)\}}. This list intentionally includes non-recyclable plastics that are found in household goods.

In addition to the pure plastic samples, we also collected commonly used plastic containers representing the aforementioned plastic types. These objects spanned diverse geometries, from cups to plastic bag fragments. This data set was made more complex by including transparent samples of these plastics, which are only present in the ``Riverbed'' scenarios and never used during training. Scenarios 7-10 also include samples of garbage that is not made of plastic, such as cardboard, newsprint, and glass. Scenarios 9-10 include pieces of fauna native to the local area of the study, including leaves and woody branches. A successful classifier should detect only the presence of plastic, and not native vegetation or non-plastic debris.
\begin{figure}[b]
\centerline{\includegraphics[trim={0 0 0 0.1cm},clip,width=\columnwidth]{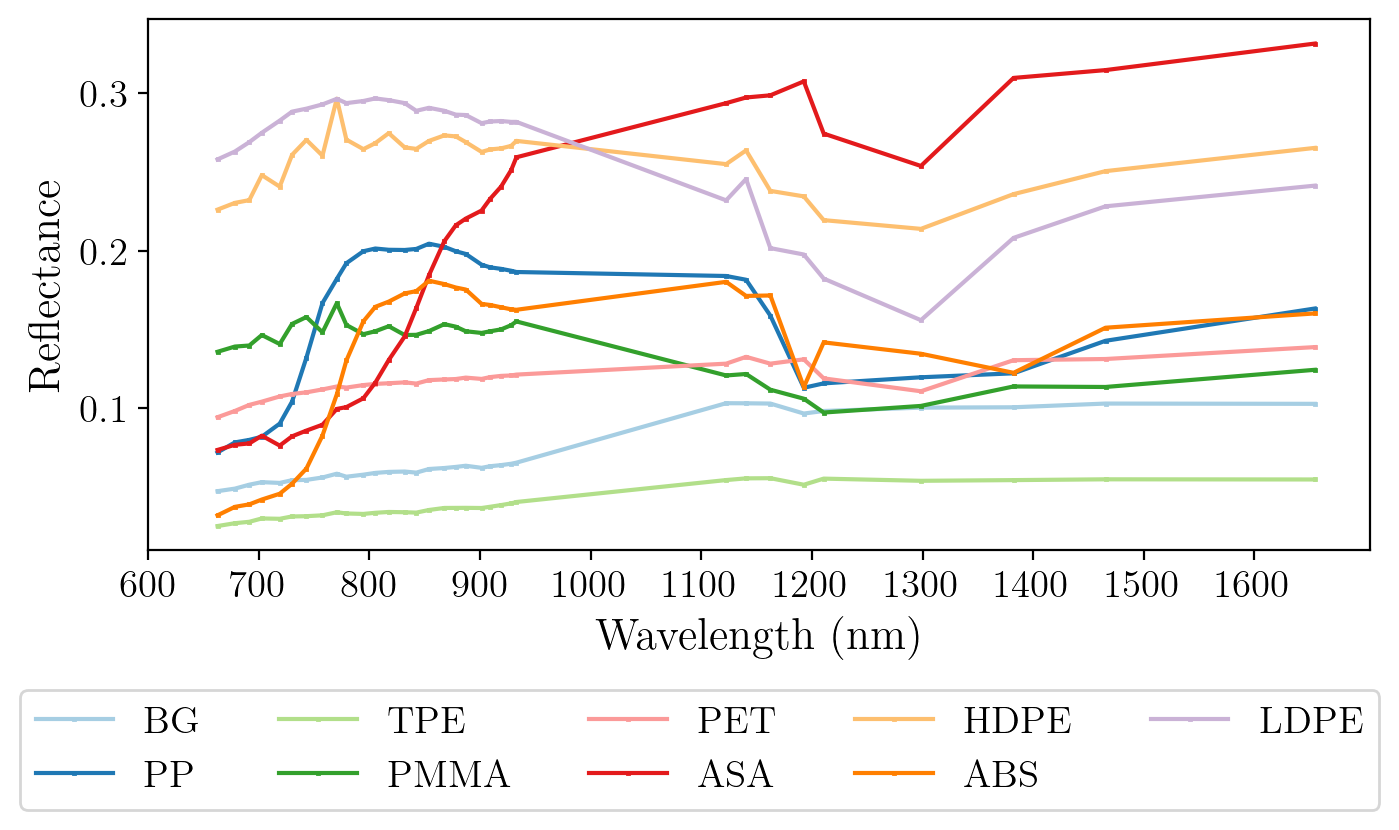}}  
\vspace{-0.2cm}
\caption{Mean spectral reflectance profiles of pure plastics.}
\label{fig:plastic_dry_spect}
\vspace{-0.5cm}
\end{figure}
\begin{table}[t]
\caption{Plastic Detection Scenarios and Composition}
\label{tab:scenario_setup}
\centering
\footnotesize
\setlength\tabcolsep{4 pt} 
\begin{tabular}{c|p{3cm}|c|c}
    \toprule
    Scenario \# & Composition & Background & Train/Test\\
    \midrule
        1 & Native Vegetation & Black PE & \cellcolor{yellow!25}Train \\ 
        2 & Plastic Samples & Black PE & \cellcolor{yellow!25}Train\\ 
        3 & Plastic Samples & Black PE, Water & \cellcolor{yellow!25}Train\\ 
        4 & Plastic Samples & Riverbed Sand & \cellcolor{yellow!25}Train\\ 
        5 & Plastic Waster & Black PE & \cellcolor{yellow!25}Train\\ 
        6 & Plastic Waste & Riverbed Sand &  \cellcolor{yellow!25}Train\\ 
        7 & Plastic Waste, Non-plastic Waste, & Settled Riverbed &  \cellcolor{green!25}Test\\ 
        8 & Plastic Waste, Non-plastic Waste, & Turbid Riverbed &  \cellcolor{green!25}Test\\ 
        9 & Plastic Waste, Non-plastic Waste, Native Vegetation & Settled Riverbed &  \cellcolor{green!25}Test\\ 
        10 & Plastic Waste, Non-plastic Waste, Native Vegetation & Turbid Riverbed & \cellcolor{green!25}Test\\ 
        \bottomrule
    \end{tabular}
    \vspace{-1.0em}
\end{table}

\subsection{Model Construction}
\vspace{-0.5em}
In this work, we leverage different classification algorithms for pixel classification, namely feed-forward neural networks (NN), logistic regression (LR), and support vector machines (SVMs), for plastic detection. All models were trained following a binary classification strategy with the two classes being $\{\mathrm{plastic},\mathrm{non\!\!-\!\!plastic}\}$.
For training and testing we used the images acquired in the scenario depicted in Table~\ref{tab:scenario_setup}. We used data from the six non-riverbed scenarios for training and validation purposes while reserving the four riverbed scenarios for testing. 
We randomly split the training data into training and validation sets with equal size. However, we actually used only half of the pixels in the training set to speed up the optimization process. Finally, we report the performance metrics obtained in the validation and test sets.

\textbf{Model details:}
All neural networks (NN) consisted of 3 hidden layers with 100, 50, and 25 units, respectively. Intermediate layers use the ReLU activation function. For NN training, we used the Adam optimizer~\cite{kingma2014adam} with a learning rate of $10^{-3}$. For SVM, we opted for a linear kernel in our experiments. While training our models, we used each pixel as an input and the corresponding label for that pixel as a target. All models were implemented with Sci-kit Learn, except for a GPU-accelerated NN implemented in PyTorch.

\section{Results}
\begin{figure*}
\centering
\includegraphics[trim={0 0 0 0.1cm}, clip,width=2\columnwidth]{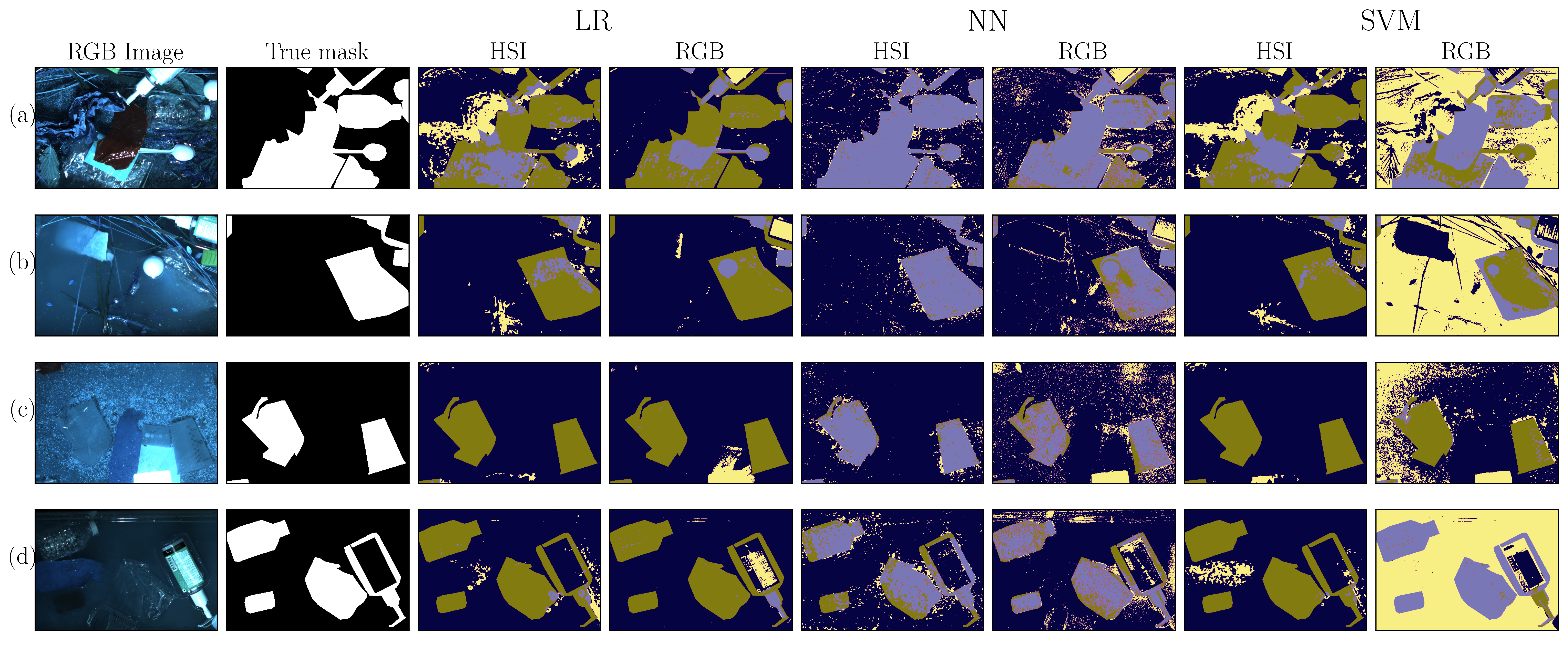} 
\vspace{-0.5cm}
\caption{Results on held-out test images. The correct detections are highlighted in light purple, while the accurate non-detections are shown in dark purple. False detections are indicated with a light yellow color, and missed detections are represented by a darker yellow shade. \textbf{(a)} Scenario \#9: Settled riverine plastics and vegetation. \textbf{(b)} Scenario \#10: Turbid riverine plastics and vegetation. \textbf{(c)} Scenario \#7: Settled riverine plastics.  \textbf{(d)} Scenario \#8: Turbid riverine plastics.}
\label{fig:binary_classification_test}
\vspace{-0.8 em}
\end{figure*}

Fig.~\ref{fig:plastic_dry_spect} depicts different spectral signatures obtained from plastic materials existing in the training sets. They were computed by averaging the reflectances of all pixels for each of the plastic components. The figure indicates that the spectral signature changes significantly between the different plastics in all spectral ranges. 

We report the performance metrics obtained with training, validation and test sets in Table~\ref{tab:results}. Specifically, we report accuracy, precision, recall, f-1 score, and Area Under the Curve (AUC) metrics for our experiments. 
We highlight that the test images are quite different from the training sets in the sense that they mimic a riverbed scenario where the plastic litter is floating or partially submerged. 

In Table \ref{tab:results}, all five metrics show that the detection performance obtained with NN is superior to those obtained with LR and SVM, in all scenarios, with NN successfully achieving classification precisions and AUC of 94\% and 98\%, respectively, when using HSI. These results demonstrate that the extra information embedded in the HSIs leads to substantial performance improvements. For example, when comparing the NN results for HSI and RGB, HSI led to a 9\% and 8\% increase in classification accuracy and AUC, respectively, for the test set. Furthermore, the 0.98 AUC achieved with NN-HSI indicates that very precise detectors can be designed with a low probability of false positives. 

Fig. \ref{fig:binary_classification_test} presents plastic detection results on the test images. These images contain various types of plastic materials, which are partially or entirely immersed in water. Notably, the utilization of NN yields more accurate detections compared to alternative models, irrespective of whether the input consists of HSI or RGB images. Particularly, the NN model trained and tested on HSI data exhibits superior performance, as substantiated by the numerical findings presented in Table \ref{tab:results}, presenting very accurate classification maps even in this simplistic pixel-wise classification setup. 
\begin{table}[bt]
\centering
\renewcommand{\arraystretch}{1}
\setlength{\tabcolsep}{3.3pt}
\caption{Detection performance of our models are demonstrated using five different metrics: Accuracy (Acc), precision (precis.), recall, f-1 score (f-1), and AUC.}

\begin{tabular}{p{0.3cm}|p{0.8cm}p{0.99cm}|p{0.9cm}p{0.9cm}p{0.9cm}p{0.9cm}p{0.9cm}}
\toprule
& Image & Model & Acc & Precis. & Recall & f-1 & AUC \\
\bottomrule
\toprule

\multirow{6}{*}{\STAB{\rotatebox[origin=c]{90}{Validation}}} &  & NN  & 0.94 & 0.94 & 0.94 & 0.94  &  0.98 \\ 
& HSI & LR  & 0.74 & 0.74 & 0.74 & 0.71  & 0.75 \\ 
& & SVM  & 0.67 & 0.66 & 0.67 & 0.62 &  0.71 \\ 
\partialmidrule{l}{2-8}
& & NN  & 0.85 & 0.86 & 0.85 & 0.85  & 0.93 \\ 
& RGB & LR  & 0.66 & 0.64 & 0.66 & 0.58 & 0.71 \\ 
& & SVM  & 0.44 & 0.53 & 0.44 & 0.44 & 0.52 \\ 
\midrule

\multirow{6}{*}{\STAB{\rotatebox[origin=c]{90}{Test}}} &  & NN  & \textbf{0.94} & \textbf{0.94} & \textbf{0.94} & \textbf{0.94} & \textbf{0.98} \\ 
& HSI & LR  & 0.73 & 0.68 & 0.73 & 0.68 & 0.64 \\ 
& & SVM  & 0.71 & 0.65 & 0.71 & 0.65 & 0.62 \\ 
\partialmidrule{l}{2-8}
& & NN  & 0.85 & 0.85 & 0.85 & 0.85  & 0.91 \\ 
& RGB & LR  &  0.74 & 0.71 & 0.74 & 0.67 & 0.65 \\ 
& & SVM  & 0.40 & 0.56 & 0.40 & 0.41 & 0.49 \\ 
\midrule

\end{tabular}
\label{tab:results}
\end{table}
\section{Discussion}
The results presented in the previous section indicate that computational imaging techniques are suitable for MPW monitoring in the riverine, leading to high classification accuracy and AUCs, especially when HSI, despite the relatively small, and irregularly placed spectral resolution (only 33 bands). For the NN with the best performance, we evaluated on both a CPU and a GPU. The inference time for a full datacube was 4000 ms and 450 ms, respectively. This speed-up near an order of magnitude shows significant promise in further optimizations for real-time evaluation of snapshot hyperspectral data.

Per-pixel classification tends to yield speckled results where erroneously classified pixels. The results in Fig.\ref{fig:binary_classification_test}b,c,and d show NNs trained on HSI minimize the speckling around the detection targets. This minimization of noise is important in discerning not only the presence of a target plastic but also its spatial size and shape. Another significant observation comes from the analysis of the false positives in the RGB images. Across all the model types trained on RGB data, there are significant false detections of plastics when the object is visibly white. Models trained with the HSI data are able to correctly identify these objects (paper) as non-plastic.

Another important finding is that transparent plastics could be detected well despite never appearing in the training images. We expected that transparent plastics could be reasonably identified when using HSI but the fact that they could be detected under the RGB setting is indeed surprising. We expect that the lack of a richer set of materials and a well-controlled laboratory setup could be among the reasons for this performance under RGB settings. The results showed that non-linear models, when compared with linear approaches such as LR and linear SVMs, can better discern plastic litter from other materials existing within the scenes.

\section{Conclusion}
In this work, we investigate the feasibility of different imaging techniques associated with machine learning algorithms for plastic litter detection in riverine. Our findings indicate that imaging-based systems are a viable solution for automatic plastic litter detection in rivers, that high accuracy can be achieved with relatively low spectral resolution hyperspectral sensors, that HSI seems more promising than RGB imaging, and that nonlinear classifiers lead to large performance gains when compared with linear ones. Future directions include performing analysis in real-time over sequences of video-rate hyperspectral data and increased complexity of the machine learning techniques. Efficient models that include both spatial and spectral characteristics and are evaluated on a GPU offer great promise. This work constitutes a promising step towards automated plastic monitoring from a deployed sensor setup with snapshot hyperspectral imaging.


\bibliographystyle{IEEEbib}
\bibliography{strings}

\end{document}